\documentclass[lettersize,journal]{IEEEtran}
\usepackage{amsmath,amsfonts}
\usepackage{algorithmic}
\usepackage{algorithm}
\usepackage{array}
\usepackage[caption=false,font=normalsize,labelfont=small,textfont=small]{subfig}
\usepackage{textcomp}
\usepackage{stfloats}
\usepackage{url}
\usepackage{verbatim}
\usepackage{graphicx}
\usepackage{cite}
\usepackage{multirow}
\usepackage{color}
\usepackage{setspace}
\usepackage{makecell}
\usepackage{pifont}
\usepackage[colorlinks,linkcolor=magenta]{hyperref}

\begin{document}



\title{YOWOv2: A Stronger yet Efficient Multi-level Detection Framework for Real-time Spatio-temporal Action Detection}

\author{Jianhua Yang$^{1,2}$, Kun Dai$^{1}$
\thanks{This work was supported in part by the National Natural Science Foundation of China (62176072).}
\thanks{$^{1}$Jianhua Yang and Kun Dai are with the State Key Laboratory of Robotics and System,
Harbin Institute of Technology, Harbin 150001, China.}
\thanks{$^{2}$Jianhua Yang is with with Wuhu Robot Industry Technology Research Institute, Harbin Institute
of Technology, Wuhu 241000, China.}}

\markboth{Journal of \LaTeX\ Class Files,~Vol.~14, No.~8, August~2021}%
{Shell \MakeLowercase{\textit{et al.}}: A Sample Article Using IEEEtran.cls for IEEE Journals}


\maketitle

\begin{abstract}
  Designing a real-time framework for the spatio-temporal action detection task is still a challenge. In this
  paper, we propose a novel real-time action detection framework, YOWOv2. In this new framework, YOWOv2 takes
  advantage of both the 3D backbone and 2D backbone for accurate action detection. A multi-level detection
  pipeline is designed to detect action instances of different scales. To achieve this goal, we carefully build
  a simple and efficient 2D backbone with a feature pyramid network to extract different levels of classification
  features and regression features. For the 3D backbone, we adopt the existing efficient 3D CNN to save development
  time. By combining 3D backbones and 2D backbones of different sizes, we design a YOWOv2 family including
  YOWOv2-Tiny, YOWOv2-Medium, and YOWOv2-Large. We also introduce the popular dynamic label assignment strategy
  and anchor-free mechanism to make the YOWOv2 consistent with the advanced model architecture design. With our
  improvement, YOWOv2 is significantly superior to YOWO, and can still keep real-time detection. Without any
  bells and whistles, YOWOv2 achieves 87.0\% frame mAP and 52.8\% video mAP with over 20 FPS on the UCF101-24. On
  the AVA, YOWOv2 achieves 21.7\% frame mAP with over 20 FPS. Our code is available on \url{https://github.com/yjh0410/YOWOv2}.
\end{abstract}

\begin{IEEEkeywords}
  Spatio-temporal action detection, one-stage detection, spatial encoder, temporal encoder
\end{IEEEkeywords}

\section{Introduction}
\label{introduction}
\IEEEPARstart{S}{patio}-temporal action detection (STAD) aims to detect action instances in the current frame.
It has been widely applied, such as video surveillance\cite{clapes2018action} and somatosensory game\cite{yan2019stat}.

Some researchers\cite{girdhar2019video, wu2020context, zhao2022tuber} employ 3D CNNs\cite{carreira2017quo, feichtenhofer2019slowfast}
to extract spatio-temporal information from video clips to accurately detect actions, as the occurrence of actions
is a continuous concept over time. However, many 3D CNN-based frameworks suffer from poor detection speeds, which
prevent them from operating in real-time due to the massive computational requirements of the hefty 3D CNNs they use.

Hence, other researchers\cite{kalogeiton2017action, song2019tacnet, li2020actions} leverage 2D CNNs\cite{liu2016ssd, zhou2019objects}
to develop more efficient action detection frameworks. The key concept behind these 2D CNN-based frameworks is
to use a parameter-sharing 2D CNN to extract spatial features frame-by-frame and store them in a buffer.
Subsequently, they only need to process the new input frame, combine its spatial features with those in the
buffer, and generate the spatio-temporal features for the final detection. Nonetheless, such a pipeline cannot
fully model temporal association, and real-time detection is only feasible with RGB streams. When optical flow
is used, although performance improves, the speed is significantly reduced.

On the contrary, K{\"o}p{\"u}kl{\"u} et al.\cite{kopuklu2019you} develops a novel one-stage action detector,
You Only Watch Once (YOWO), by combining a 2D backbone\cite{redmon2017yolo9000} for spatial localization
and a 3D backbone for spatio-temporal modeling. To mitigate the high computational cost of 3D CNNs, they
designs a series of efficient 3D CNNs\cite{kopuklu2019resource} as the 3D backbone for efficient inference.
After the backbones, YOWO employs a channel encoder to fuse the two features for the final detection. With
their designs, YOWO achieves excellent performance on popular benchmarks and is touted as a fast action detector.
However, YOWO still suffers from two disadvantages:

\begin{itemize}
  \item YOWO is a one-level detector and performs the final detection on a low-level feature map, impairing the
  detection performance for small action instances.
  \item YOWO is an anchor-based method and has lots of anchor boxes with many hyperparameters, such as the number,
  size, and aspect ratio of anchor boxes. Those hyperparameters must be carefully artificially designed, impairing
  the generalization.
\end{itemize}

In summary, \textbf{designing a real-time detection framework for spatio-temporal action detection remains a challenge}.

In this study, we propose YOWOv2, a brand-new real-time action detector. A 3D backbone with a multi-level 2D
backbone make up YOWOv2. A multi-level detection pipeline is designed for YOWOv2 to detect action occurrences of
various scales thanks to our multi-level 2D backbone with a feature pyramid network (FPN)\cite{lin2017feature}.
We also recommend the quick deployment 3D CNNs\cite{kopuklu2019resource} for the 3D backbone. The disadvantages
of the anchor box are also avoided by using the anchor-free mechanism. We use a dynamic label assignment
technique because the anchor box is removed, enhancing the adaptability of the YOWOv2. Moreover, we construct a
variety of YOWOv2 models, such as \textbf{YOWOv2-Tiny}, \textbf{YOWOv2-Medium} and \textbf{YOWOv2-Large} by merging
3D backbones with 2D backbones of various sizes for platforms with diverse computing power.

Compared to YOWO, YOWOv2 delivers superior performance on the UCF101-24\cite{soomro2012ucf101} and AVA\cite{gu2018ava}
datasets and boasts significant advantages in terms of both parameter count and FLOPs. Moreover, YOWOv2 is
capable of real-time operation. In comparison to other real-time action detectors, YOWOv2 also achieves better
performance. In summary, our contributions are as follows:

\begin{itemize}
  \item We propose a new real-time action detection framework, YOWOv2 with a multi-level detection structure,
  which is friendly to detect small action instances.
  \item YOWOv2 features an anchor-free detection pipeline, which eliminates the limitations of anchor boxes. 
  \item We design a YOWOv2 family by combining the 3D backbones and 2D backbones of different sizes for the
  platforms with different computing power.
  \item YOWOv2 achieves state-of-the-art performance on popular benchmarks, compared to other real-time
  action detectors. 
\end{itemize}

\begin{figure*}[]
  \centering
  \includegraphics[width=0.95\linewidth]{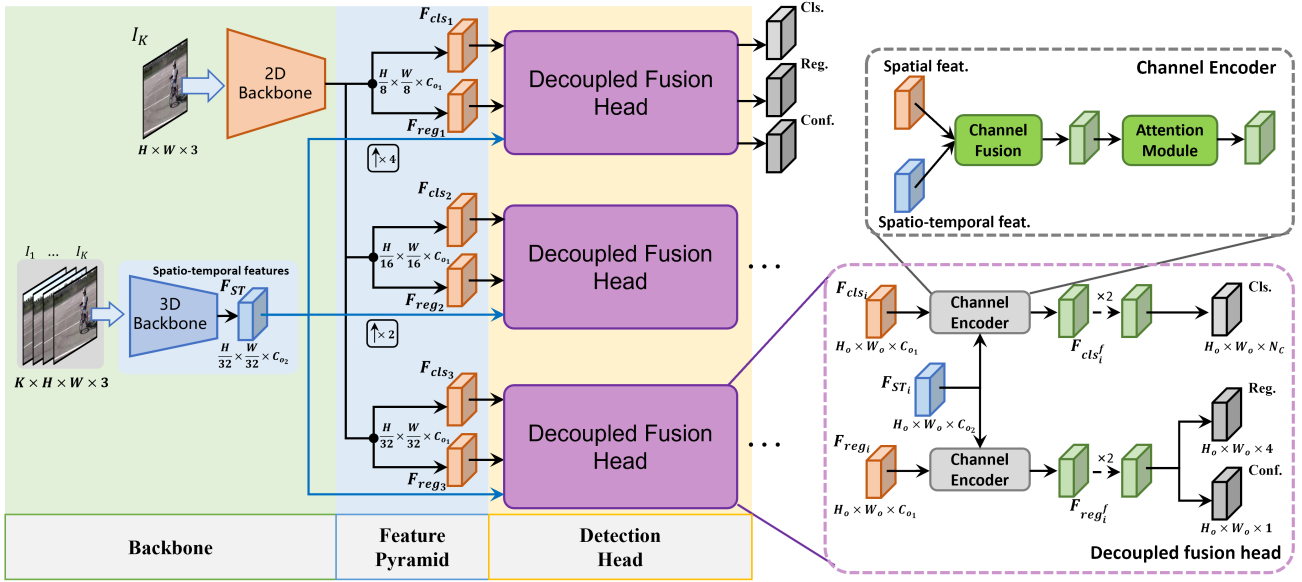}
  \caption{Overview of YOWOv2. YOWOv2 uses upsampling operation to align the spatio-temporal features output
  by the 3D backbone with the spatial features of each level output by the 2D bakcbone and uses the Decoupled
  fusion head to achieve the fusion of the two features on each level. Finally, YOWOv2 outputs the multi-level
  confidence predictions, classification predictions, and regression predictions respectively.}
  \label{fig:yowo_v2_net}
\end{figure*}

\section{Related work}
\label{related_work}
\subsection{Spatio-temporal action detection}
Spatio-temporal action detection involves detecting and identifying all instances of action that occur within
a given frame. To achieve accurate action detection, it is essential to effectively extract spatio-temporal features.

\textbf{3D CNN-based methods}. Some researchers use the 3D CNN to design action detectors\cite{hou2017tube, duarte2018videocapsulenet, girdhar2019video, wu2020context, chen2021watch, pan2021actor},
due to the strong spatio-temporal modeling capabilities. Girdhar et al.\cite{girdhar2019video} use the
I3D\cite{carreira2017quo} to generate action region proposals and then use the Transformer\cite{vaswani2017attention}
to complete the final detection. Zhao et al.\cite{zhao2022tuber} deploy a 3D CNN to encode input video and
then use the Transformer with the tuber queries for final detection. Although these 3D CNN-based methods
achieve impressive success, they all suffer from the expensive computation of the heavy 3D CNN and are therefore 
too slow to run in real time.

\textbf{2D CNN-based methods}. Another approach is to separate spatio-temporal associations and design 2D CNN-based
action detectors for efficient detection. For instance, Kalogeiton et al.\cite{kalogeiton2017action} devise
a one-stage detection framework called ActionTubelet (ACT). They utilize SSD\cite{liu2016ssd} to extract spatial
features from each frame in a video clip and then merge them. Subsequently, a detection head is employed to
process the merged spatial features for the final detection. Li et al.\cite{li2020actions} follow the ACT framework
and develop an anchor-free one-stage action detector called MovingCenter (MOC). Ma et al.\cite{ma2021spatio}
further enhance the MOC with a self-attention mechanism. However, the real-time detection performance of these
methods can only be ensured when RGB streams are applied as input. When optical flow is added, their speed
significantly declines, despite the improved performance. Moreover, obtaining high-quality optical flow requires
offline processing, which cannot meet the demands of online operations.


\section{Methodology}
\label{method}
\subsection{Preliminary}\label{subsection:preliminary}
The overview of YOWOv2 is shown in Fig.\ref{fig:yowo_v2_net}. Given a video clip with $K$ frames
$V=\{I_1, I_2, \dots, I_K \}$ where the $I_K$ is the current frame, YOWOv2 uses an efficient 3D CNN\cite{kopuklu2019resource}
as the 3D backbone to extract spatio-temporal features $F_{ST} \in \mathbb{R}^{\frac{H}{32}\times \frac{H}{32}\times C_{o_2}}$.
The 2D backbone of YOWOv2 is a multi-level 2D CNN, responsible for outputting decoupled multi-level spatial
features $F_{cls} = \{ F_{cls_i} \}_{i=1}^{3}$ and $F_{reg} = \{ F_{reg_i} \}_{i=1}^{3}$ of $I_K$, where the
$F_{cls_i} \in \mathbb{R}^{\frac{H}{2^{i+2}}\times \frac{W}{2^{i+2}}\times C_{o_1}}$ is the classification
features and $F_{reg_i} \in \mathbb{R}^{\frac{H}{2^{i+2}}\times \frac{W}{2^{i+2}}\times C_{o_1}}$ is the
regression features. After the two backbones, we deploy two channel encoders on each feature map of level to
integrate features. After that, two extra parallel branches with two $3\times 3$ conv layers followed the
channel encoders to predict $Y_{cls_i} \in \mathbb{R}^{\frac{H}{2^{i+2}}\times \frac{W}{2^{i+2}}\times N_C}$
for classification, $Y_{reg_i} \in \mathbb{R}^{\frac{H}{2^{i+2}}\times \frac{W}{2^{i+2}}\times 4}$ for
regression respectively. A confidence branch is added on the regression branch to predict
$Y_{conf_i} \in \mathbb{R}^{\frac{H}{2^{i+2}}\times \frac{W}{2^{i+2}}\times 1}$ for actionness confidence.
Next, we introduce the design of YOWOv2 in detail.

\subsection{Design of YOWOv2}
\label{subsection:yowov2_design}
\textbf{2D backbone}.
The 2D backbone is supposed to extract multi-level spatial features from the current frame. Considering the
balance between performance and speed, we draw some advanced ideas from the advanced object detectors\cite{wang2022yolov7, ge2021yolox}.
We reuse the backbone and feature pyramid network (FPN) of YOLOv7\cite{wang2022yolov7} to save training time.
After the FPN, we add extra $1\times 1$ conv layers to compress the channel number of each level feature map
$F_{S_i}$ to $C_{o_1}$ which is defaulted to 256. Then, we add two parallel branches with two $3\times 3$
conv layers to output decoupled features, as shown in Eq.(\ref{eq:decoupled_feat}).
\begin{equation}
  \begin{aligned}
  F_{cls_i} &= f_{conv_2}^{1}\left(f_{conv_1}^{1}\left(F_{S_i} \right) \right) \\
  F_{reg_i} &= f_{conv_2}^{2}\left(f_{conv_1}^{2}\left(F_{S_i} \right) \right) \\
  \end{aligned}
  \label{eq:decoupled_feat}
\end{equation}
where the $f_{conv_j}^{i}$ is the $j^{th}$ $3\times 3$ conv layer of the $i^{th}$ branch.

In YOWOv2 framework, the 2D backbone outputs the decoupled feature maps of three levels, $F_{cls} = \{ F_{cls_i} \}_{i=1}^{3}$
and $F_{reg} = \{ F_{reg_i} \}_{i=1}^{3}$. We name the 2D backbone FreeYOLO for convenience. By controlling
the depth and width of FreeYOLO, we designed two FreeYOLO of different sizes, FreeYOLO-Tiny for YOWOv2-Tiny
and FreeYOLO-Large for YOWOv2-Medium and YOWOv2-Large. To accelerate the convergence of training, we pretrain
our 2D backbone with additional $1\times 1$ conv layers on the COCO\cite{lin2014microsoft}. The pretrained
weight files are available on the GitHub\footnote[1]{\url{https://github.com/yjh0410/FreeYOLO}}.

\begin{figure}[]
  \centering
  \includegraphics[width=1.0\linewidth]{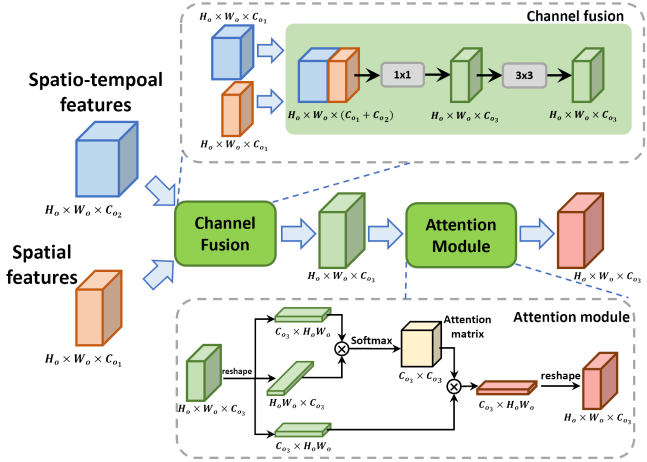}
  \caption{Overview of ChannelEncoder. It contains the channel fusion and channel self-attention mechanism,
  which are both used to fuse 2D and 3D features.}
  \label{fig:channel_encoder}
\end{figure}

\textbf{3D backbone}.
The 3D backbone is supposed to extract the spatio-temporal features $F_{ST}$ from the video clip for the
spatio-temporal association. We deploy the efficient 3D CNN\cite{kopuklu2019resource} to reduce computation
and thus guarantee real-time detection. To fuse with decoupled spatial features, we simply upsample $F_{ST}$
to obtain $\{ F_{ST_i} \}_{i=1}^{3}$, as shown in Eq.(\ref{eq:upsample_3d}).
\begin{equation}
  \begin{aligned}
  F_{ST_1} &= \mathtt{Upsample}_{4\times}\left(F_{ST} \right) \\
  F_{ST_2} &= \mathtt{Upsample}_{2\times}\left(F_{ST} \right) \\
  F_{ST_3} &= F_{ST} \\
  \end{aligned}
  \label{eq:upsample_3d}
\end{equation}
where the $\mathtt{Upsample}$ is the upsampling operation for aligning
$F_{ST_i}\in \mathbb{R}^{\frac{H}{2^{i+2}}\times \frac{H}{2^{i+2}}\times C_{o_2}}$ and $F_{cls_i}$ and
$F_{reg_i}$ in the spatial dimension.

\textbf{ChannelEncoder}.
ChannelEncoder, proposed by YOWO\cite{kopuklu2019you}, is supposed to fuse the features from the 2D backbone
and 3D backbone. Given a $F_S \in \mathbb{R}^{H_o\times W_o\times C_{o_1}}$ and a $F_{ST} \in \mathbb{R}^{H_o\times W_o\times C_{o_2}}$,
the ChannelEncoder first concatenates them along the channel dimension and uses two naive conv layer followed
a BN and LeakyReLU to achieve primary channel integration, as following,
\begin{equation}
  F_{f} =  f_{conv_2}\left(f_{conv_1}\left(\mathtt{Concat}\left[F_{S}, F_{ST} \right]\right)\right)
\end{equation}
where the $F_f \in \mathbb{R}^{H_o\times W_o\times C_{o_3}}$, $\mathtt{Concat}$ is the channel concatenation
operation, $f_{conv_1}$ and $f_{conv_2}$ are both the conv layers with BN and LeakyReLU. Then, the $F_f$ is
reshaped to $F_{f_2} \in \mathbb{R}^{C_{o_3}\times H_oW_o}$ for the following channel self-attention
mechanism inspired by DANet\cite{fu2019dual} to do deeper processing, so that the information containing
two different levels features can be fully integrated, as shown in Eq.(\ref{eq:channel_self_attn})
\begin{equation}
  F_{f_3} = \mathtt{Softmax}\left( F_{f_2} F_{f_2}^{T} \right)F_{f_2}
  \label{eq:channel_self_attn}
\end{equation}

Finally, the $F_{f_3} \in \mathbb{R}^{C_{o_3}\times H_oW_o}$ is reshaped to $F_{f} \in \mathbb{R}^{H_o\times W_o\times C_{o_3}}$
followed by another conv layer. The whole pipeline of the ChannelEncoder is shown in Fig.\ref{fig:channel_encoder}.

\textbf{Decoupled fusion head}.
In YOWOv2, the 2D backbone outputs the decoupled spatial features $F_{cls} = \{ F_{cls_i} \}_{i=1}^{3}$ and
$F_{reg} = \{ F_{reg_i} \}_{i=1}^{3}$ of the current frame $I_K$ while the 3D backbone outputs $\{ F_{ST_i} \}_{i=1}^{3}$
obtained by upsampling $F_{ST}$ of the video clip $V=\{I_1, I_2, \dots, I_K \}$. Note that $F_{cls_i}$ and
$F_{reg_i}$ contain very different semantic information, which inspires us to fuse $F_{cls_i}$ and $F_{reg_i}$
with $F_{ST_i}$ separately. Therefore, we design a decoupled fusion head to fuse $F_{ST_i}$ into $F_{cls_i}$
and $F_{reg_i}$ independently, as shown in Eq.(\ref{eq:decoupled_head}).
\begin{equation}
  \begin{aligned}
  F_{cls_i}^{f} &=  \mathtt{ChannelEncoder}\left(F_{cls_i}, F_{ST_i} \right) \\
  F_{reg_i}^{f} &=  \mathtt{ChannelEncoder}\left(F_{reg_i}, F_{ST_i} \right)
  \end{aligned}
  \label{eq:decoupled_head}
\end{equation}

After the feature aggregation, we deploy two parallel branches on each level for final detection. Its design
is simple, just a classification branch and a box regression branch. 

For the classification branch, it outputs the classification prediction $Y_{cls_i} \in \mathbb{R}^{\frac{H}{2^{i+2}}\times \frac{W}{2^{i+2}}\times N_C}$,
where $Y_{cls_i}(x,y)$ represents the probability of action instances at each spatial position on $Y_{cls_i}$
and $N_C$ is the number of action classes. Taking $F_{cls_i}^{f}$, the branch applies two $3 \times 3$ conv layers,
each with $C$ filters and each followed by SiLU activations. Finally, a $1 \times 1$ conv layer with $N_C$
filters and sigmoid activations is attached to output the $N_C$ binary predictions per spatial position.

For the box regression branch, it outputs the box regression prediction $Y_{reg_i} \in \mathbb{R}^{\frac{H}{2^{i+2}}\times \frac{W}{2^{i+2}}\times 4}$,
where $Y_{reg_i}(x,y)$ represents the 4 relative offsets at each spatial position. The design is equal to the
classification branch except that the final $1 \times 1$ conv layer is with $4$ filters for offset predictions.
Additionally, an extra $1 \times 1$ conv layer with $1$ filter is added into this branch for actionness confidence
prediction, $Y_{conf_i} \in \mathbb{R}^{\frac{H}{2^{i+2}}\times \frac{W}{2^{i+2}}\times 1}$. Note that, there
is no anchor box in each spatial position, therefore, YOWOv2 is an anchor-free method. 

\subsection{Label assignment}
\label{subsection:label_assignment}
Since YOWOv2 is an anchor-free action detector without any anchor boxes, the multi-level label assignment becomes
a challenge. Recently, dynamic label assignment has shown success in object detection. Inspired by YOLOX\cite{ge2021yolox},
we implement SimOTA for the label assignment of YOWOv2. Specifically, we calculate the cost between all predicted
bounding boxes and groundtruths. Eq.(\ref{eq:cost}) demonstrates the cost between the $i^{th}$ prediction and
the $j^{th}$ ground truth. Subsequently, each groundtruth is assigned with the $top_k$ predicted bounding boxes
with the least cost, where $k$ is dynamically determined by the IoU between the predicted bounding boxes and the
target bounding boxes.

\begin{equation}
  \begin{aligned}
    c_{ij}\left( \hat{a}_i, a_j, \hat{b}_i, b_j \right) & = L_{cls}(\hat{a}_i, a_j) + \gamma L_{seg}(\hat{b}_i, b_j) \\
  \end{aligned}
  \label{eq:cost}
\end{equation}
where the $\hat{a}_i$ and $a_j$ are the classification prediction (multiplied by confidence prediction) and target,
$\hat{b}_i$ and $b_j$ are the regression prediction and target and $\gamma$ is the cost balance factor, empirically
being 3 in the experiments.

\subsection{Loss function}
\label{subsection:loss}
We define loss function as follows:
\begin{equation}
  \begin{aligned}
    L(\{a_{x,y}\}, \{b_{x,y}\}, \{c_{x,y}\}) &= \frac{1}{N_{pos}}\sum_{x,y}L_{conf}(\hat{c}_{x,y}, c_{x,y}) \\
    & + \frac{1}{N_{pos}}\sum_{x,y}\mathbb I_{\{\hat{a}_{x,y} > 0\}}L_{cls}(\hat{a}_{x,y}, a_{x,y}) \\
    & + \frac{\lambda}{N_{pos}} \sum_{x,y}\mathbb I_{\{\hat{a}_{x,y} > 0\}}L_{reg}(\hat{b}_{x,y}, b_{x,y})
  \end{aligned}\label{eq:loss}
  \end{equation}
where $L_{conf}$ and $L_{cls}$ are both the binary cross-entropy and $L_{reg}$ is the GIoU loss\cite{rezatofighi2019generalized}.
The $a_{x,y}$, $b_{x,y}$ and $c_{x,y}$ are classification prediction, regression prediction, and confidence prediction,
while the $\hat{a}_{x,y}$, $\hat{b}_{x,y}$ and $\hat{c}_{x,y}$ are groundtruths. $N_{pos}$ denotes the number of
positive samples and $\lambda$ is the loss balance factor, empirically being 5 in the experiments.
$I_{\{\hat{a}_{x,y} > 0\}}$ is the indicator function, being 1 if $\hat{a}_{x,y} > 0$ and 0 otherwise.

\section{Experiments}\label{experiments}
\subsection{Datasets}
\label{subsection:datasets}
\textbf{UCF101-24}\cite{soomro2012ucf101}. UCF101-24 contains 3,207 untrimmed videos for 24 sports classes
and provides corresponding spatio-temporal annotations. There may be multiple action instances per frame. Following
YOWO\cite{kopuklu2019you}, we train and evaluate YOWO-Plus on the first split.

\textbf{AVA}\cite{gu2018ava}. AVA is a large-scale benchmark for spatial-temporal action detection. It contains
430 15-minute video clips with 80 atomic visual actions (AVA). It provides annotations at 1 Hz in space and
time, and precise spatio-temporal annotations with possibly multiple annotations for each person. Therefore,
this benchmark is very challenging. We train YOWOv2 on the train split and evaluate it on the most-frequent
60 action classes of the AVA dataset. We report evaluation results on the AVA v2.2.

\begin{figure}[]
  \centering
  \includegraphics[width=0.90\linewidth]{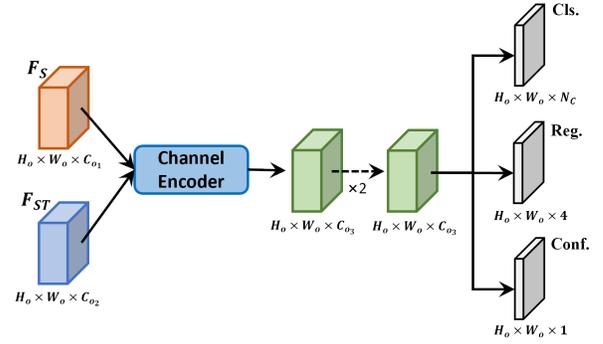}
  \label{fig:coupled_head}
  \caption{Coupled fusion head. In the coupled head, the spatial features from the 2D backbone is also coupled
  which means that the parallel $3\times 3$ conv layers after the FPN are removed.}
  \label{fig:cfh}
\end{figure}

\begin{table}[]
  \caption{Performance comparison between Coupled fusion head (CFH) and Decoupled fusion head (DFH) on the UCF101-24.}
  \centering
  \resizebox{1.0\linewidth}{!}{
    \begin{tabular}{c|c|c|c|c}
      \hline
      Head                 & Model    & FPS    & F-mAP (\%)    & V-mAP (\%)     \\ \hline
      \multirow{3}{*}{CFH} & YOWOv2-T & 56    &  78.9          &  49.8          \\
                           & YOWOv2-M & 45    &  81.2          &  50.7          \\
                           & YOWOv2-L & 33    &  84.3          &  51.5          \\ \hline
      \multirow{3}{*}{DFH} & YOWOv2-T & 50    &  80.5          &  51.3          \\
                           & YOWOv2-M & 42    &  83.1          &  50.7          \\
                           & YOWOv2-L & 30    &  \textbf{85.2} &  \textbf{52.0} \\ \hline
    \end{tabular}}
    \label{comparison_ch_dh}
\end{table}

\subsection{Implementation details}
\label{subsection:implement}
For training, we use the AdamW optimizer with an initial learning rate 0.0001 and weight decay 0.0005. The batch
size is set to 8 with 16 gradient accumulate. On the UCF101-24, we train YOWOv2 for 7 epochs and decay the
learning rate by a factor of 2 at 1, 2, 3, and 4 epoch, respectively. On the AVA, we train YOWOv2 for 9 epochs
and decay the learning rate by a factor of 2 at 3, 4, 5, and 6 epoch, respectively. Unless otherwise specified,
the size of the input frame is reshaped to $224 \times 224$.

For evaluation metrics, we follow previous works\cite{kopuklu2019you, chen2021watch, zhao2022tuber} to report
frame mAP (F-mAP) and video mAP (V-mAP) at 0.5 IoU between predictions and groundtruths. We follow the link
algorithm of YOWO\cite{kopuklu2019you} to build action tubelets. On the AVA, we report frame mAP at 0.5 IoU
since the annotations are sparsely provided at 1 Hz.

\subsection{Effectiveness of decoupled fusion head}
\label{subsection:ablation_study}
To evaluate the impact of the decoupled fusion head on YOWOv2, we design a coupled fusion head as a control group,
as shown in Fig.\ref{fig:cfh}. We conduct experiments on UCF101-24 and the results are summarized in Table\ref{comparison_ch_dh}.
The table shows that the decoupled fusion head outperforms the coupled fusion head. These results indicate
that feature fusion should be performed decoupled due to the semantic differences between categorical and
regressive features. Although the decoupled fusion head slightly slows down the detection speed, the significant
improvement in performance compensates for the marginal loss in speed.

\begin{table}[]
  \caption{Ablation study of the effectiveness of loss balance factor $\lambda$.}
  \centering
  \resizebox{0.8\linewidth}{!}{
  \begin{tabular}{c|c|c|c}
  \hline
  \multirow{2}{*}{$\lambda$} & \multicolumn{2}{c|}{UCF101-24}               & AVA       \\ \cline{2-4} 
                             & \multicolumn{1}{c|}{F-mAP (\%)} & V-mAP (\%) & mAP (\%)  \\ \hline
  1.0                        & 83.3                            & 50.1       & 19.6      \\
  2.0                        & 84.7                            & 51.1       & 19.8      \\
  3.0                        & 85.1                            & 51.9       & 20.0      \\
  4.0                        & 85.2                            & 52.0       & 20.2      \\
  5.0                        & 85.2                            & 52.0       & 20.2      \\
  6.0                        & 85.0                            & 52.0       & 20.1      \\
  7.0                        & 84.8                            & 51.8       & 20.0      \\ \hline
\end{tabular}}
\label{loss_lambda}
\end{table}

\subsection{Effectiveness of the loss balance factor}
\label{subsection:ablation_study_lambda}
We also verify the effect of loss balance factor $\lambda$ defined in Eq.(\ref{eq:loss}). Table.\ref{loss_lambda}
summarizes the results on the UCF101-24 and AVA. From the table, YOWOv2 is insensitive to $\lambda$ in the range
3 to 6, but the larger or smaller $\lambda$  weakens the performance of the YOWOv2. Therefore, we set $\lambda$
to 5 in the experiments.

\begin{table}[]
  \caption{Comparison with YOWO on the UCF101-24. FPS is measured on a GPU RTX 3090. K is the length of the video clip.}
  \centering
  \resizebox{1.0\linewidth}{!}{
  \begin{tabular}{c|c|c|c|c|c|c}
  \hline
  Method     & K  & FPS   & F-mAP (\%)  & V-mAP (\%)  & GFLOPs     & Params      \\ \hline
  YOWO       & 16 & 34    & 80.4        & 48.8        & 43.7       & 121.4 M     \\ \hline
  YOWOv2-T   & 16 & 50    & 80.5        & 51.3        & 2.9        & 10.9 M      \\
  YOWOv2-M   & 16 & 42    & 83.1        & 50.7        & 12.0       & 52.0 M      \\
  YOWOv2-L   & 16 & 30    & 85.2        & 52.0        & 53.6       & 109.7 M     \\ \hline
  YOWOv2-T   & 32 & 50    & 83.0        & 51.2        & 4.5        & 10.9 M      \\
  YOWOv2-M   & 32 & 40    & 83.7        & 52.5        & 12.7       & 52.0 M      \\
  YOWOv2-L   & 32 & 22    & 87.0        & 52.8        & 91.9       & 109.7 M      \\ \hline

  \end{tabular}}
  \label{comparison_yowo_ucf24}
\end{table}
  
\begin{table}[]
  \caption{Comparison with YOWO on the AVA. FPS is measured on a GPU RTX 3090.}
  \centering
  \begin{tabular}{c|c|c|c|c}
    \hline
    Method      & K  & FPS   & mAP   & GFLOPs  \\ \hline
    YOWO        & 16 & 31    & 17.9  & 44      \\
    YOWO        & 32 & 23    & 19.1  & 82      \\
    YOWO+LFB    & -  & -     & 20.2  & -       \\ \hline
    YOWOv2-T    & 16 & 49    & 14.9  & 3       \\
    YOWOv2-M    & 16 & 41    & 18.4  & 12      \\
    YOWOv2-L    & 16 & 29    & 20.2  & 54      \\ \hline
    YOWOv2-T    & 32 & 49    & 15.6  & 5       \\
    YOWOv2-M    & 32 & 40    & 18.4  & 13      \\
    YOWOv2-L    & 32 & 22    & 21.7  & 92      \\ \hline
\end{tabular}
\label{comparison_yowo_ava}
\end{table}

\subsection{Comparison with YOWO}
\label{subsection:comparison_yowo}
\begin{figure}[]
  \centering
  \includegraphics[width=1.0\linewidth]{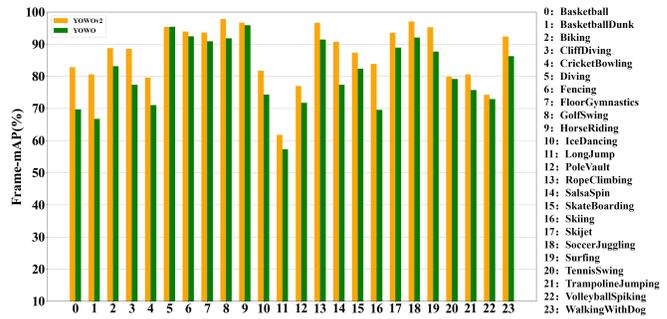}
  \caption{Per-class frame mAP at 0.5 IoU on the UCF101-24. The orange bars represent the per-class AP of YOWOv2-L, while
           the green bars represent the per-class AP of YOWO.}
  \label{fig:per_cls_ap}
\end{figure}

To compare the accuracy, speed, and computation of YOWOv2 with YOWO\cite{kopuklu2019you}, we design three scales
of YOWOv2 by combining different 3D backbones and 2D backbones: YOWOv2-Tiny (YOWOv2-T), YOWOv2-Medium (YOWOv2-M),
and YOWOv2-Large (YOWOv2-L). To demonstrate the superior speed and performance balance of YOWOv2, we compare our
YOWOv2 family with YOWO on the UCF101-24. The comparison results are summarized in Table.\ref{comparison_yowo_ucf24}.
The table shows that YOWOv2-T outperforms YOWO in terms of both frame mAP (80.5 \% v.s. 80.4 \%) and video mAP
(51.3 \% v.s. 48.8 \%) with significantly fewer FLOPs (2.9G vs. 43.7G) and parameters (10.9M vs. 121.4M), while
achieving higher FPS (50 vs. 34) on an RTX 3090 GPU. Moreover, with a stronger 2D backbone and 3D backbone, YOWOv2-L
achieves the best performance. Fig.\ref{fig:per_cls_ap} shows the per-class AP comparison results between YOWO
and YOWOv2-L.

\begin{figure}[]
  \centering
  \includegraphics[width=1.0\linewidth]{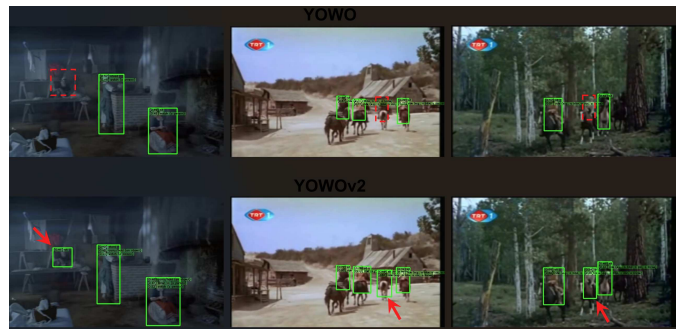}
  \caption{Performance comparison of small action instance detection between YOWO and YOWOv2 on the AVA.}
  \label{fig:compare_small_act}
\end{figure}

We also conduct a comparative experiment with YOWO on the AVA benchmark. The comparison results are summarised in
Table.\ref{comparison_yowo_ava}. It is unrealistic to expect the little YOWOv2-T to outperform YOWO, which has
greater calculations and more parameters, given that the AVA is a highly difficult dataset. Here, the YOWOv2-L is
what we focus on most. YOWOv2-L accomplishes a superior trade-off between performance and detection speed as compared
to YOWO. YOWOv2-L performs better than YOWO with the LFB as well. The benefits on these well-known benchmarks
demonstrate that YOWOv2's design is superior to that of YOWO, meeting the goals of inheritance and development and
generating a new generation of real-time action detection framework.

On the other hand, to illustrate the effectiveness of YOWOv2's multi-level detection, we compare the performance
of small action instance detection with YOWO, as shown in Fig.\ref{fig:compare_small_act} The figure shows that
YOWO misses certain smaller action instances because it has insufficient confidence in them (red dotted line boxes).
YOWOv2 can more effectively detect the tiny action instances that YOWO cannot handle since it is equipped with the
multi-level detection pipeline.

\begin{figure}[]
  \centering
  \includegraphics[width=1.0\linewidth]{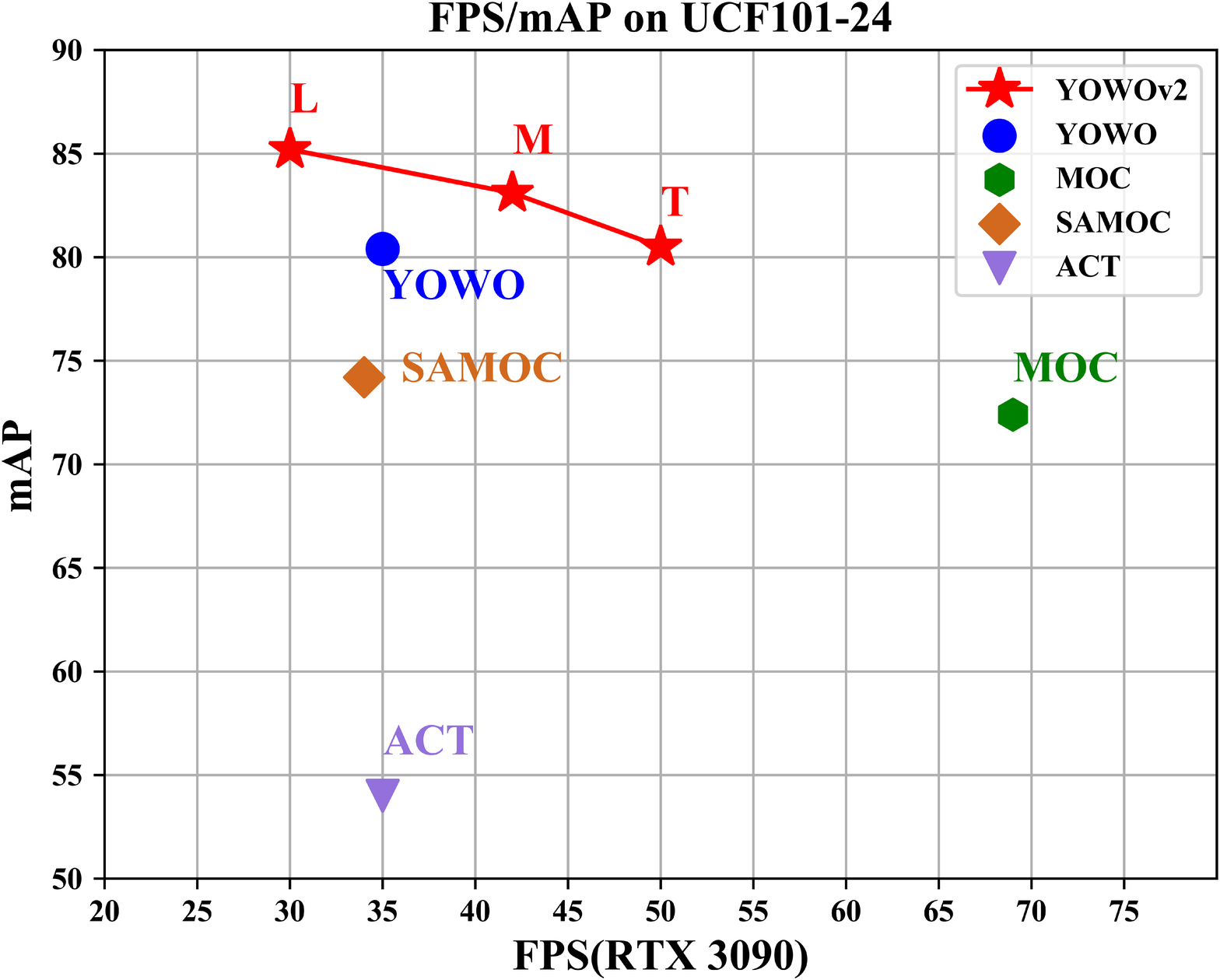}
  \caption{Speed/accuracy trade-off among multiple real-time action detectors, including YOWO, MOC, SAMOC, ACT
           and the proposed YOWOv2. Speed is measured on an NVIDIA 3090 GPU with batch size 1. Note that the length
           of the input video clip is 16 YOWO and YOWOv2.}
  \label{fig:fps_map}
\end{figure}

\subsection{Comparison with other real-time action detectors}
\label{subsection:comparison_real_time}
We contrast YOWOv2 with additional real-time action detectors in addition to YOWO. Fig.\ref{fig:fps_map} shows
the speed/accuracy trade-off of those detectors that can run at over 25 FPS on an RTX 3090 GPU, including YOWO\cite{kopuklu2019you},
MOC\cite{li2020actions}, SAMOC\cite{ma2021spatio} and ACT\cite{kalogeiton2017action}. As shown in the figure,
YOWOv2 greatly improves on the performance and detection speed trade-off. YOWOv2 can be seen as a new generation
of superior real-time motion detectors as a result.

\subsection{Comparison with state-of-the-art works}
\label{subsection:comparison_sota}
\textbf{UCF101-24}.
Table.\ref{comparison_ucf24} summarizes the comparison results with state-of-the-art works on the UCF101-24.
For stronger performance, we also use $K=32$ to train and test YOWOv2. The majority of 2D CNN-based detectors
extract richer spatio-temporal characteristics from the optical flow in parallel with the video clip to improve
their performance. Unfortunately, using optical flow not only reduces the model's applicability because it's
challenging to get high-quality optical flow online in real-time, but it also slows down the speed of detection.
Our real-time YOWOv2 still performs admirably when measured against the potent 3D CNN-based techniques.


\begin{table}[]
  \caption{Comparison with state-of-the-art works on the UCF101-24. We report Frame mAP at 0.5 IoU and Video mAP at 0.5
  IoU on the first split.}
  \centering
  \resizebox{1.0\linewidth}{!}{
  \begin{tabular}{cl|c|c|c|c}
  \hline
  \multicolumn{2}{c|}{Method}                                                              & RGB         & Flow        & F-mAP (\%)    & V-mAP (\%)     \\ \hline
  \multicolumn{1}{c|}{\multirow{4}{*}{3D}}           & T-CNN\cite{hou2017tube}             & \checkmark  & \ding{56}   & 41.4          & -              \\ 
  \multicolumn{1}{c|}{}                              & I3D\cite{gu2018ava}                 & \checkmark  & \checkmark  & 76.6          & \textbf{59.9}  \\ 
  \multicolumn{1}{c|}{}                              & Tuber\cite{zhao2022tuber}           & \checkmark  & \ding{56}   & 83.2          & 58.4           \\ \hline
  \multicolumn{1}{c|}{\multirow{6}{*}{2D}} & ACT\cite{kalogeiton2017action}      & \checkmark  & \checkmark  & 67.1          & 51.4           \\ 
  \multicolumn{1}{c|}{}                              & TACNet\cite{song2019tacnet}         & \checkmark  & \checkmark  & 72.1          & 54.4           \\ 
  \multicolumn{1}{c|}{}                              & MOC\cite{li2020actions}             & \checkmark  & \ding{56}   & 73.1          & 51.0           \\ 
  \multicolumn{1}{c|}{}                              & MOC\cite{li2020actions}             & \checkmark  & \checkmark  & 78.0          & 53.8           \\ 
  \multicolumn{1}{c|}{}                              & SAMOC\cite{ma2021spatio}            & \checkmark  & \ding{56}   & 74.2          & 49.8           \\ 
  \multicolumn{1}{c|}{}                              & SAMOC\cite{ma2021spatio}            & \checkmark  & \checkmark  & 79.3          & 52.5           \\ \hline
  \multicolumn{1}{c|}{}                              & YOWOv2-T                            & \checkmark  & \ding{56}   & 80.5          & 51.3           \\
  \multicolumn{1}{c|}{}                              & YOWOv2-M                            & \checkmark  & \ding{56}   & 83.1          & 50.7           \\
  \multicolumn{1}{c|}{}                              & YOWOv2-L                            & \checkmark  & \ding{56}   & 85.2          & 52.0           \\ \hline
  \multicolumn{1}{c|}{}                              & YOWOv2-T (K=32)                     & \checkmark  & \ding{56}   & 83.0          & 51.2           \\
  \multicolumn{1}{c|}{}                              & YOWOv2-M (K=32)                     & \checkmark  & \ding{56}   & 83.7          & 52.5           \\
  \multicolumn{1}{c|}{}                              & YOWOv2-L (K=32)                     & \checkmark  & \ding{56}   & \textbf{87.0} & 52.8           \\ \hline

  \end{tabular}}
\label{comparison_ucf24}
\end{table}

\textbf{AVA}.
Table \ref{comparison_ava} summarizes the comparison results on the AVA. Since the AVA is a very challenging benchmark
where the data scene is changeable, and each action instance is labeled with multiple annotations, most current works
take advantage of the 3D CNN to challenge this dataset. Since the FLOPs of these 3D CNN-based detectors are too high
to run in real-time, we attribute them to \textit{Non real-time spatio-temporal action detectors}.

From the table, Tuber is a state-of-the-art action detector with the highest mAP on the AVA. However, its GFLOPs is as
high as 120 and detection speed is as low as 3 FPS, although it achieves 31.7 \% mAP. While having excellent performance,
such a sluggish detector is exceedingly difficult to use in practical situations. Contrary to Tuber, we currently place
more emphasis on the practicalities, specifically the GFLOPs and the FPS, despite the fact that the mAP metric is quite
vital. Although YOWOv2 has a lower mAP than Tuber, its detection speed can satisfy real-time requirements (over 20 FPS),
making it possible to use it in real-world situations to complete tasks.

\begin{figure*}[]
  \centering
  \includegraphics[width=1.0\linewidth]{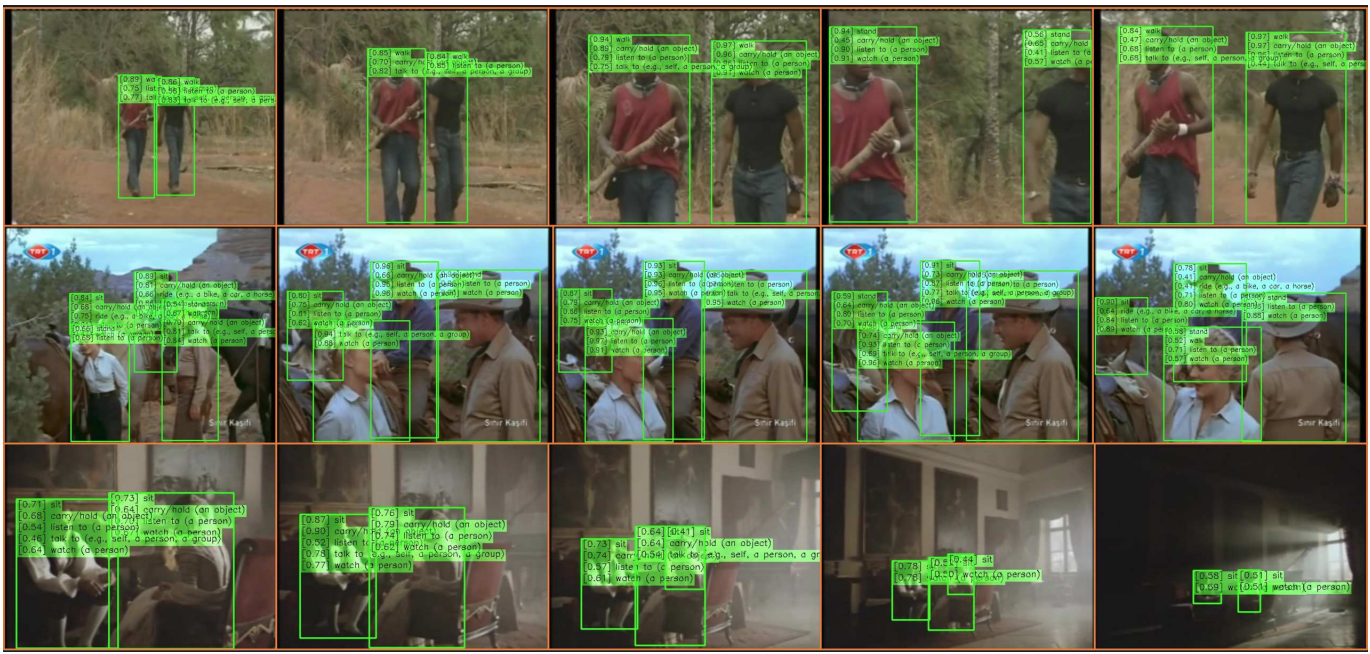}
  \caption{Qualitative results on the AVA.}
  \label{fig:vis_ava}
\end{figure*}

Fig.\ref{fig:vis_ava} shows some qualitative results on the AVA. From the figure, we can see that YOWOv2 can accurately
detect the basic postures, such as \textbf{walk} and \textbf{stand} and other actions of each person. This result shows
that YOWOv2 has the ability to understand multiple behaviors that occur in a person, which is helpful for in-depth
understanding of human behavioral intentions in the future.

\begin{table*}[]
  \caption{Comparison with state-of-the-art works on the AVA. We report mAP at 0.5 IoU on the AVA validation split. K is
  the length of the input video clip. "-" indicates that the model code has not been available, so that the speed cannot
  be tested.}
  \centering
  \begin{tabular}{c|c|c|c|c|c}
    \hline
    Method                                   & 3D Backbone          & K  & FPS & mAP (\%)  & GFLOPs    \\ \hline
    \multicolumn{6}{l}{\textit{Non real-time spatio-temporal action detector}}                                                                     \\ \hline
    WOO\cite{chen2021watch}                  & SlowFast-R101        & 8  &  -  & 28.3      & 252       \\
    SE-STAD\cite{sui2023simple}              & SlowFast-R101        & 8  &  -  & 29.3      & 165       \\
    Tuber\cite{zhao2022tuber}                & CSN-50               & 32 & 10  & 28.6      & 78       \\
    Tuber\cite{zhao2022tuber}                & CSN-152              & 32 &  3  & 31.7      & 120       \\ \hline
    \multicolumn{6}{l}{\textit{Real-time spatio-temporal action detector}}                                                                         \\ \hline
    YOWO\cite{kopuklu2019you}                & 3D-ResNeXt-101       & 16 & 35  & 17.9      & 44        \\
    YOWO\cite{kopuklu2019you}                & 3D-ResNeXt-101       & 32 & 25  & 19.1      & 82        \\ \hline
    YOWOv2-T                                 & 3D-ShuffleNetv2-1.0x & 16 & 50  & 14.9      & 2.9       \\
    YOWOv2-M                                 & 3D-ShuffleNetv2-2.0x & 16 & 42  & 18.4      & 12.1      \\
    YOWOv2-L                                 & 3D-ResNeXt-101       & 16 & 30  & 20.3      & 53.6     \\ \hline
    YOWOv2-T (K=32)                          & 3D-ShuffleNetv2-1.0x & 32 & 50  & 15.6      & 4.5       \\
    YOWOv2-M (K=32)                          & 3D-ShuffleNetv2-2.0x & 32 & 40  & 18.4      & 13.7      \\
    YOWOv2-L (K=32)                          & 3D-ResNeXt-101       & 32 & 22  & 21.7      & 92.0     \\ \hline
    \end{tabular}
    \label{comparison_ava}
  \end{table*}

\subsection{Test in real scenarios}
\begin{figure*}[]
  \centering
  \includegraphics[width=1.0\linewidth]{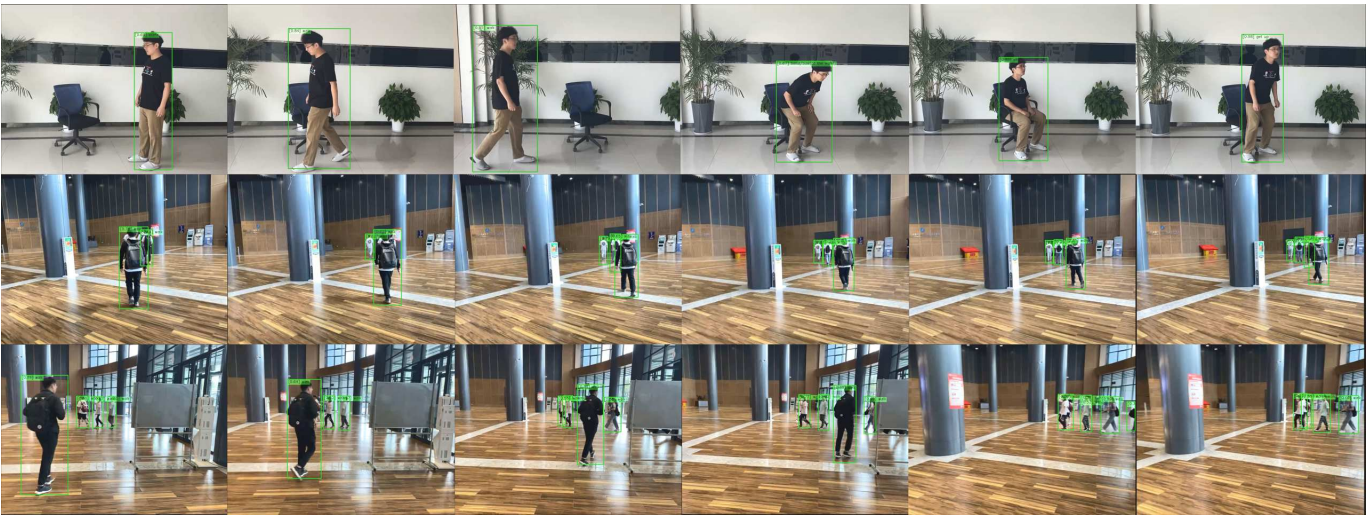}
  \caption{Qualitative results on real scene. Since most of the atomic actions in the AVA dataset do not appear
  in our real scenes, we only show fourteen basic action poses, including bend or bow, crawl, stand, walk, sit, etc.
  The green bounding box represents the spatial localization. The action category with confidence score is shown
  in the upper left corner of the bounding box.}
  \label{fig:real_scene}
\end{figure*}

To demonstrate the generalization of YOWOv2, we also test the performance of YOWOv2 in real scenarios. Fig.\ref{fig:real_scene}
shows a demo of YOWOv2 in a real scene. The input frame is reshaped to $224 \times 224$, following the requirements in
Sec.\ref{subsection:implement}. Since most of the atomic actions in the AVA dataset do not appear in our real scenes,
we only show fourteen basic action poses\cite{gu2018ava} in Fig.\ref{fig:real_scene}, including bend or bow, crawl, stand,
walk, sit, etc. From the figure, we can see that YOWOv2 still works well in real scenarios, demonstrating its effectiveness
and generalization.

\section{Conclusion}
\label{subsection:conclusion}
In this paper, we propose a novel real-time detection framework YOWOv2 for spatial-temporal action detection. The
YOWOv2 family contains YOWOv2-Tiny, YOWOv2-Medium, and YOWOv2-Large for the platforms with different computing power.
Compared to the previous version of YOWO, our YOWOv2 is designed as a multi-level action detection framework,
helping to detect smaller motion instances. YOWOv2 is also an anchor-free action detector, avoiding the drawbacks
of anchor boxes existing in YOWO. On the popular benchmarks, YOWOv2 significantly outperforms YOWO and other real-time
action detectors with a large gap. Even compared with powerful but no speed advantage 3D CNN-based methods, YOWOv2
still shows competitive performance. YOWOv2 is an effective attempt, but it is not the end of YOWO. In the future,
we will further study to design a more efficient and powerful feature pyramid network to fuse multi-level features
from both the 3D backbone and 2D backbone, not just the 2D backbone.

\bibliographystyle{ieeetr}
\bibliography{egbib.bib}

\vfill

\end{document}